# Implementation of Licensed Plate Detection and Noise Removal in Image Processing

**GAO YIQUAN**

*BSc (Hons) in Intelligent Systems, Asia Pacific University, Malaysia*

## ABSTRACT

Car license plate recognition system is an image processing technology used to identify vehicles by capturing their Car License Plates. The car license plate recognition technology is also known as automatic number-plate recognition, automatic vehicle identification, car license plate recognition or optical character recognition for cars. In Malaysia, as the number of vehicle is increasing rapidly nowadays, a pretty great number of vehicle on the road has brought about the considerable demands of car license plate recognition system. Car license plate recognition system can be implemented in electronic parking payment system, highway toll-fee system, traffic surveillance system and as police enforcement tools. Additionally, car license plate recognition system technology also has potential to be combined with various techniques in other different fields like biology, aerospace and so on to achieve the goal of solving some specialized problems.

## I.INTRODUCTION

Car license plate recognition is classified into two significant parts which are respectively plate detection and character recognition, consisting of capturing the image of car, extracting the image of license plate, extracting the characters from license plate image and at the end, recognizing license plate characters. In this project, what part the developer is going to focus on and work on is the plate detection. According to the methods and algorithms the developer plans and utilizes to reach the goal of the project, the plate detection is mainly including the following essential proposed methods which respectively are image conversion from RGB to gray scale, image equalization, blur filtering, edge detection, dilation, segmentation and noise removal [1].

## II.OBJECTIVES

The goal of the project is to achieve the plate detection, which means to locate and extract the plate from the image at the end by applying various proposed methods in order to remove all the noises of the image.

The objectives can be as below:

- To enhance the car plate recognition.
- To assist police in finding the stolen cars.
- To limit down the speeding cars.

## III.PROBLEM DOMAIN

### 3.1 UNBALANCED INTENSITY DISTRIBUTION ON THE HISTOGRAM

After the conversion of image from RGB to gray scale, the gray image always has an unbalanced intensity distribution on the histogram as shown in the left of “Fig.1”, since the conversion may increase the contrast of background noise, while decreasing the usable signal. As far as can be seen from the gray image in “Fig.2”, unbalanced intensity distribution on the histogram usually will have some bad influences on the representation of usable data like the center part containing the plate in the image, which results in some difficulties for the machine to analyze and recognize the goal [2].

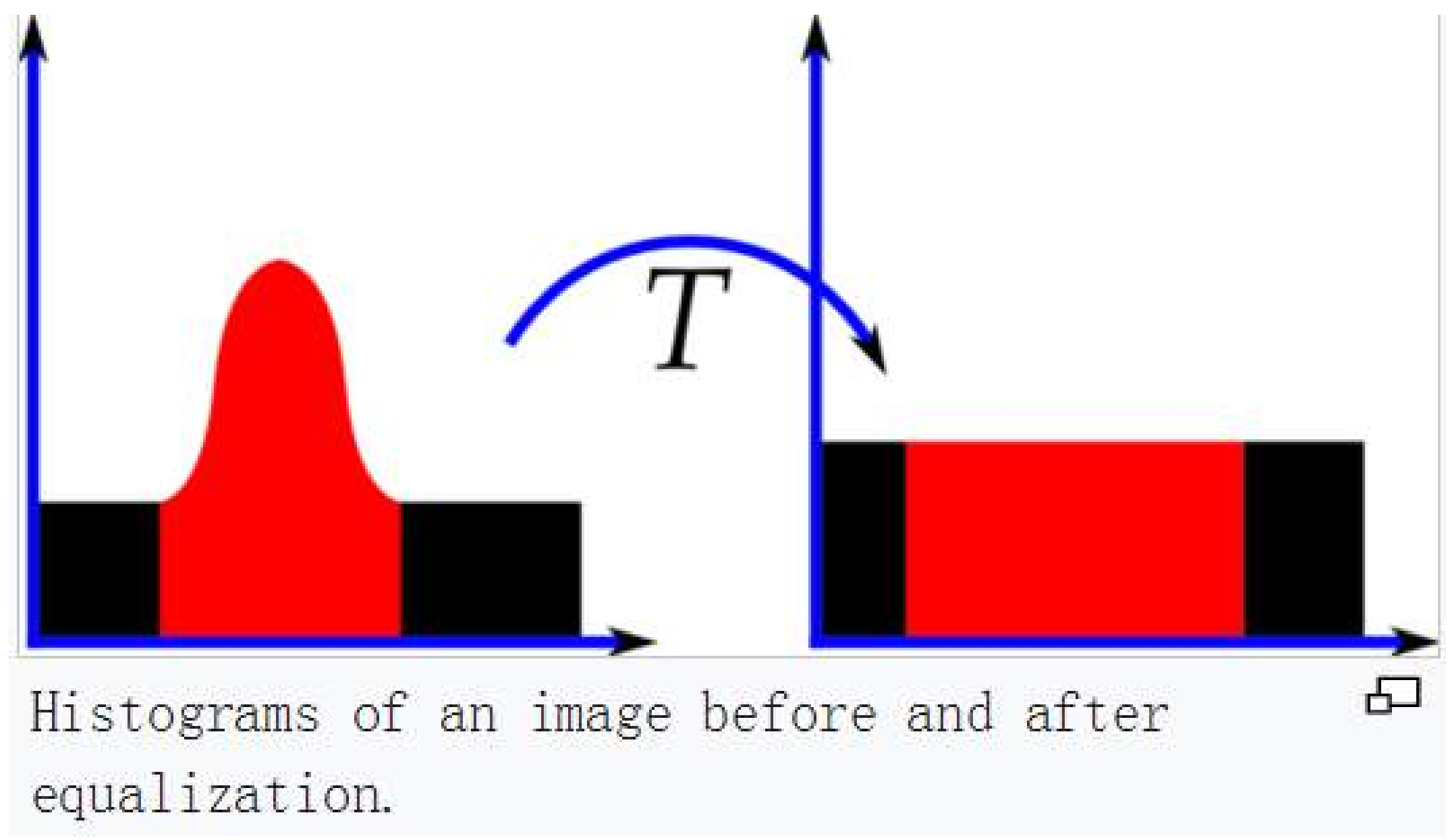

Figure.1Histogram [5]

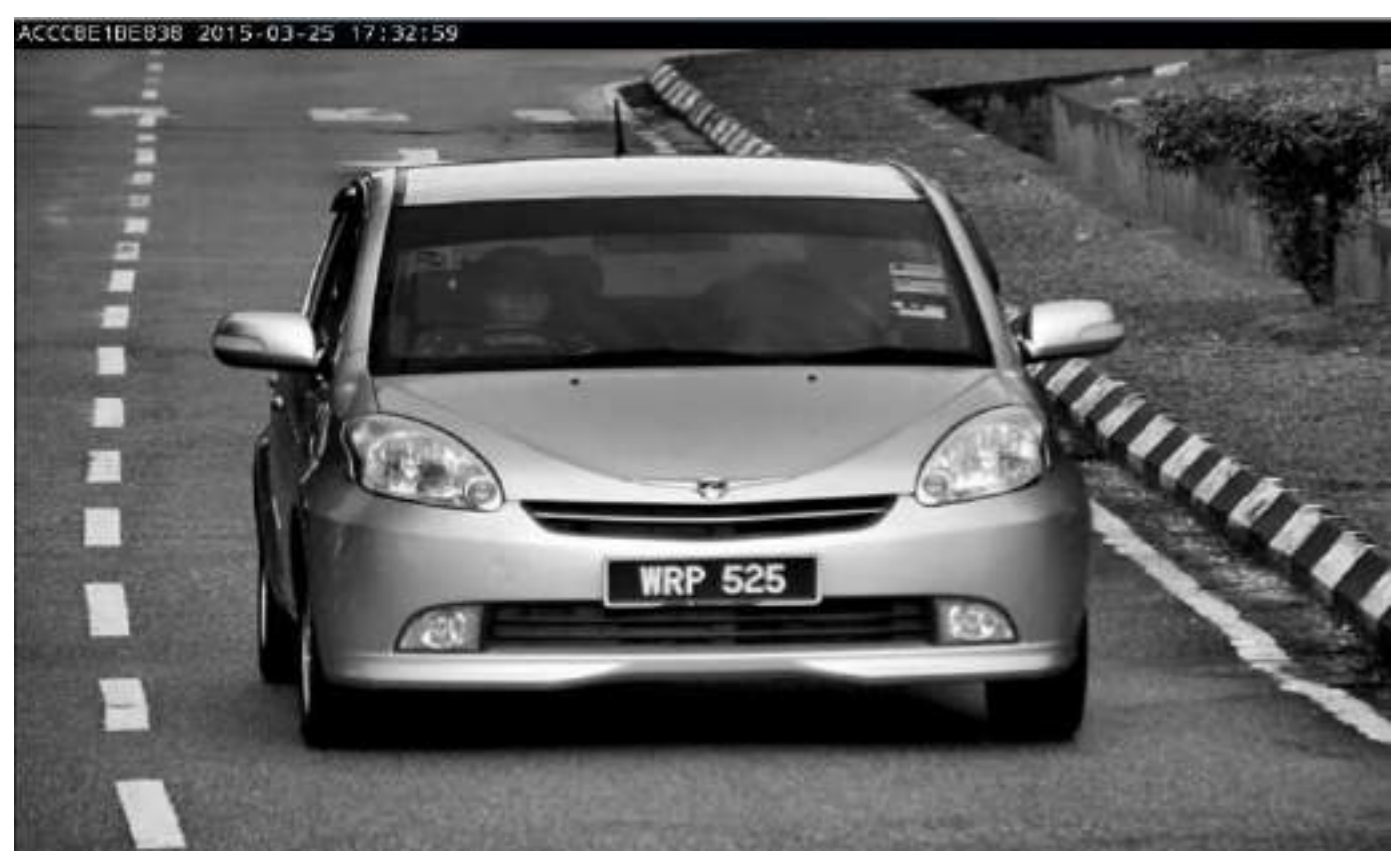

Figure.2 Gray Image

## 3.2 IMAGE NOISE

In the field of image processing, noise is always an existing problem which appears in every image waiting for processing and enhancement. Whether noise from background or foreground, both of them will obstruct the location of the plate, which leads to lower accuracy of the plate detection.

## 3.3 OUTLINE OBTAINMENT OF PLATE

Even after removing a lot of noise from background and foreground, the plate is still inside the environment with a large number of irrelevant objects surrounding it, which is quite a great obstacle to further determining the position of plate. Thus, having some tries to gain the outline of plate from the environment as well as remove irrelevant objects as many as possible is pretty of significance.

## 3.4 CONNECTION OF PIXELS INTO BLOCKS

After identifying the edge for objects, there is a problem that each object has only a fuzzy shape consisting of scattered pixels as shown in “Fig.3”. It is of importance to connect all the relevant pixels together in order to make each object a complete block, since it is more convenient and accurate to find the plate from blocks than to find the plate from scattered pixels.

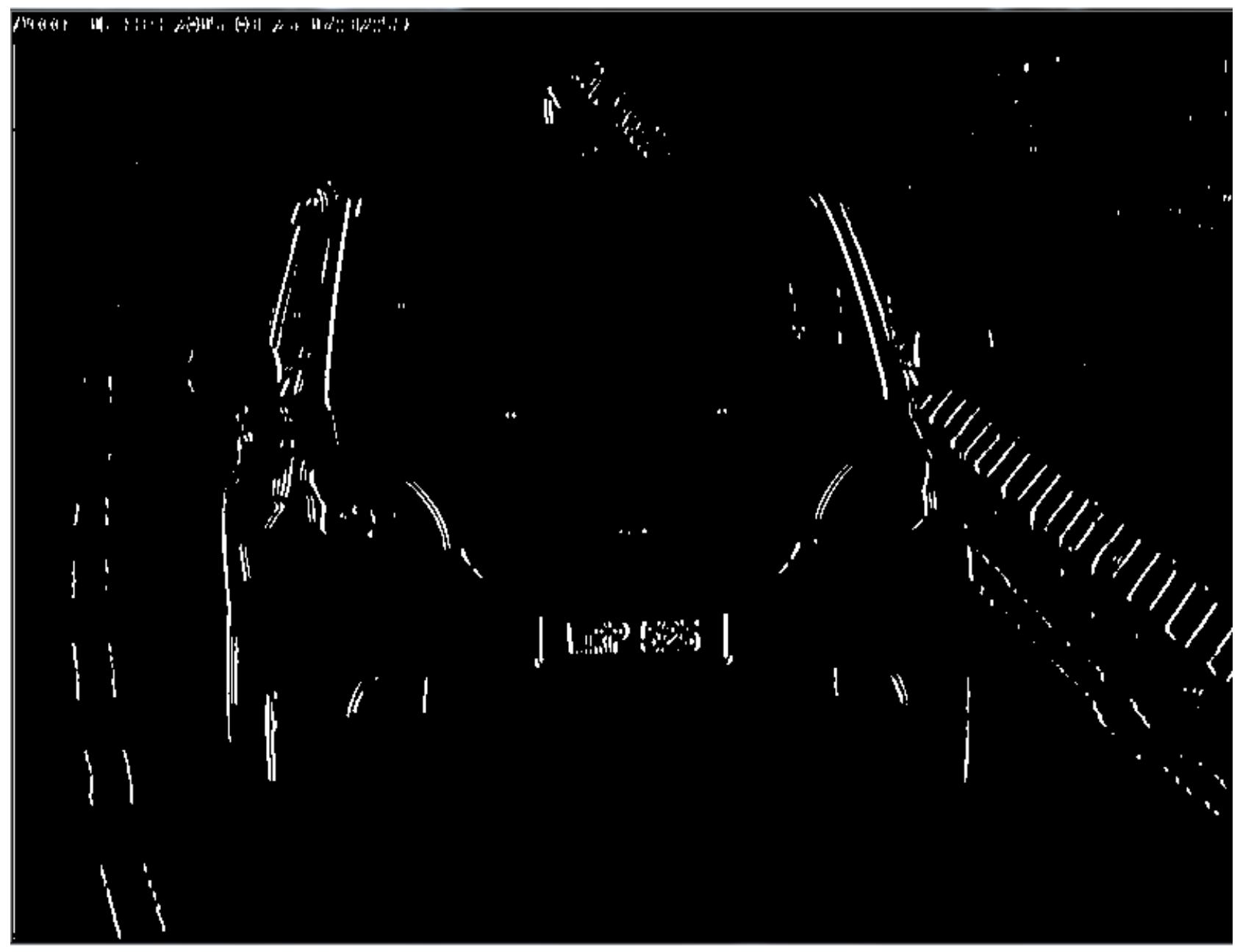

Figure.3Edge

## 3.5 INDIVIDUAL PROCESSING OF EACH BLOCK

After making different groups of pixels connected as different blocks, all the blocks are still integrated together in one image as shown in “Fig.4”. For detecting and filtering different individual blocks, it is important to find some ways to separate them as each unit, since it is much easier and more accurate to process each unit of block than to process the whole blocks together.

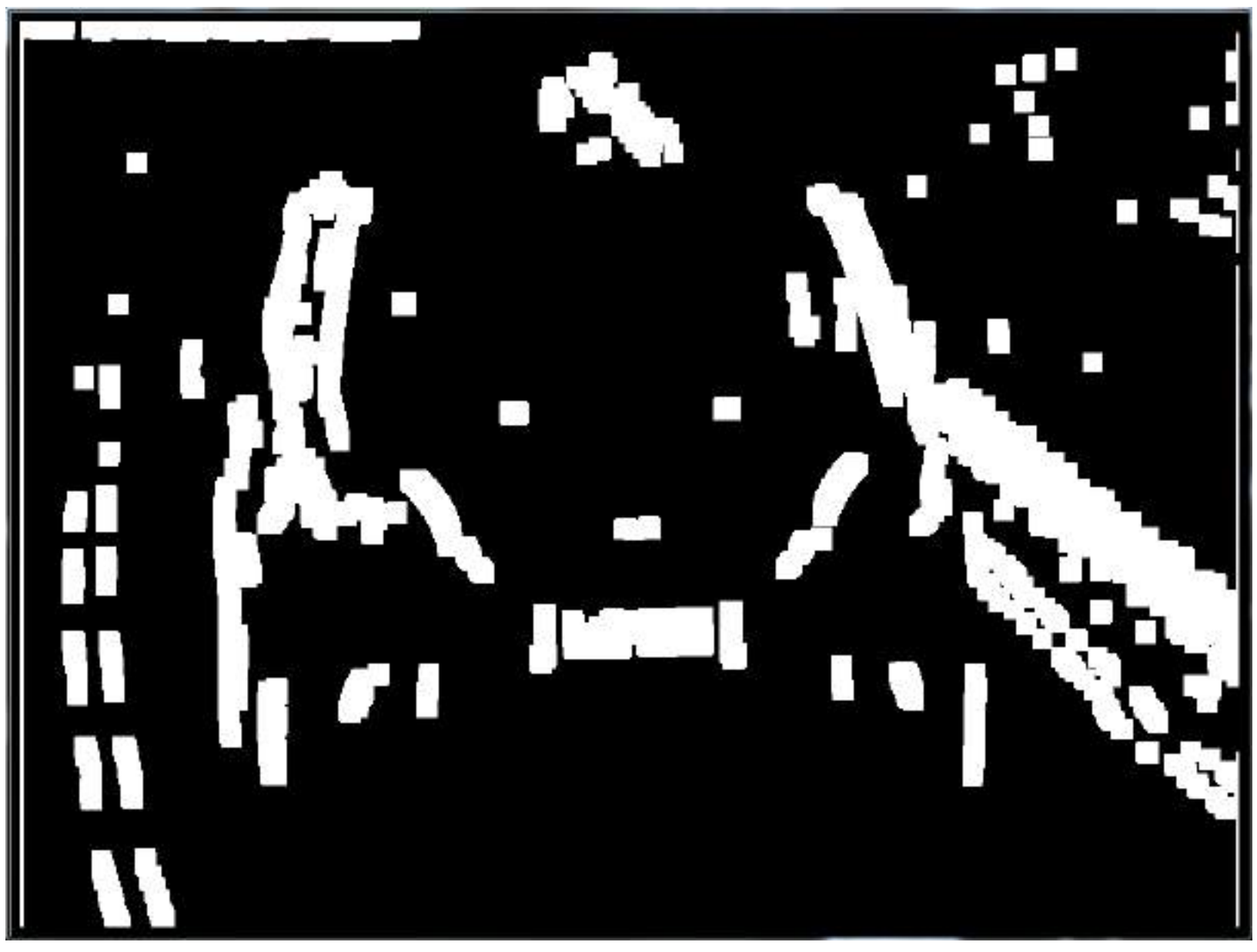

Figure.4 Integrated blocks

### 3.6 PLATE FILTRATION CHALLENGE AMONG SIMILAR BLOCKS

Although the plate filtration methods relying on heuristic features like locations, width, height and ratio are able to filter out most of the irrelevant blocks, sometimes still a small number of blocks left after filtration as the noises against the plate location and extraction. The type of problem usually is due to the similarities between the plate and those noise objects. For example, since the car logo has the similar contrast of intensity with the plate in the equalized gray scale image as “Fig.5” shows and after connecting each group of pixels, the blocks of them have the close heights and widths like “Fig.6”, at the end, the developer is faced with an unexpected result after utilizing the plate filtration methods, as shown in “Fig.7” still with two blocks left as plate candidates.

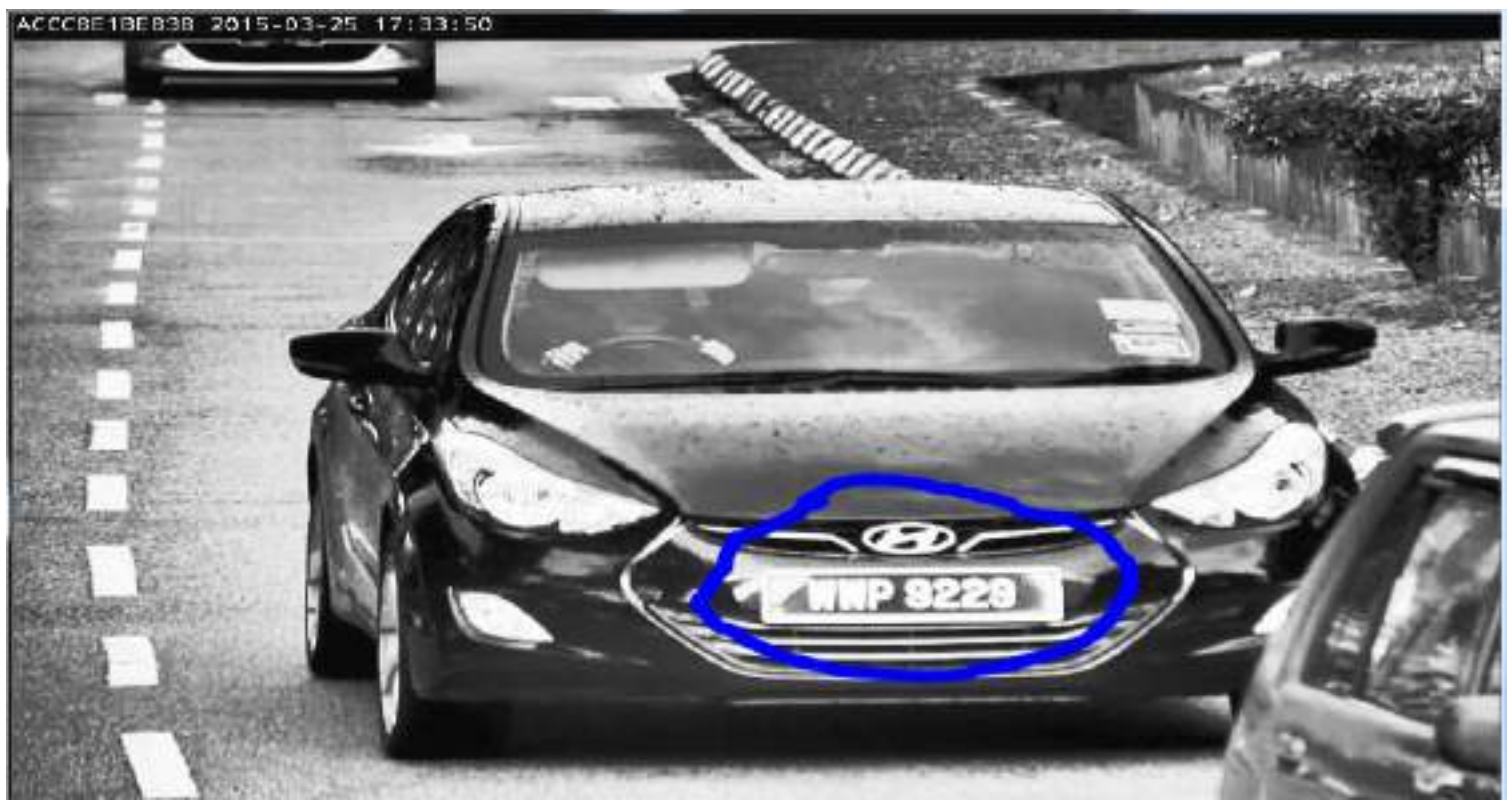

Figure.5Equalized gray scale image

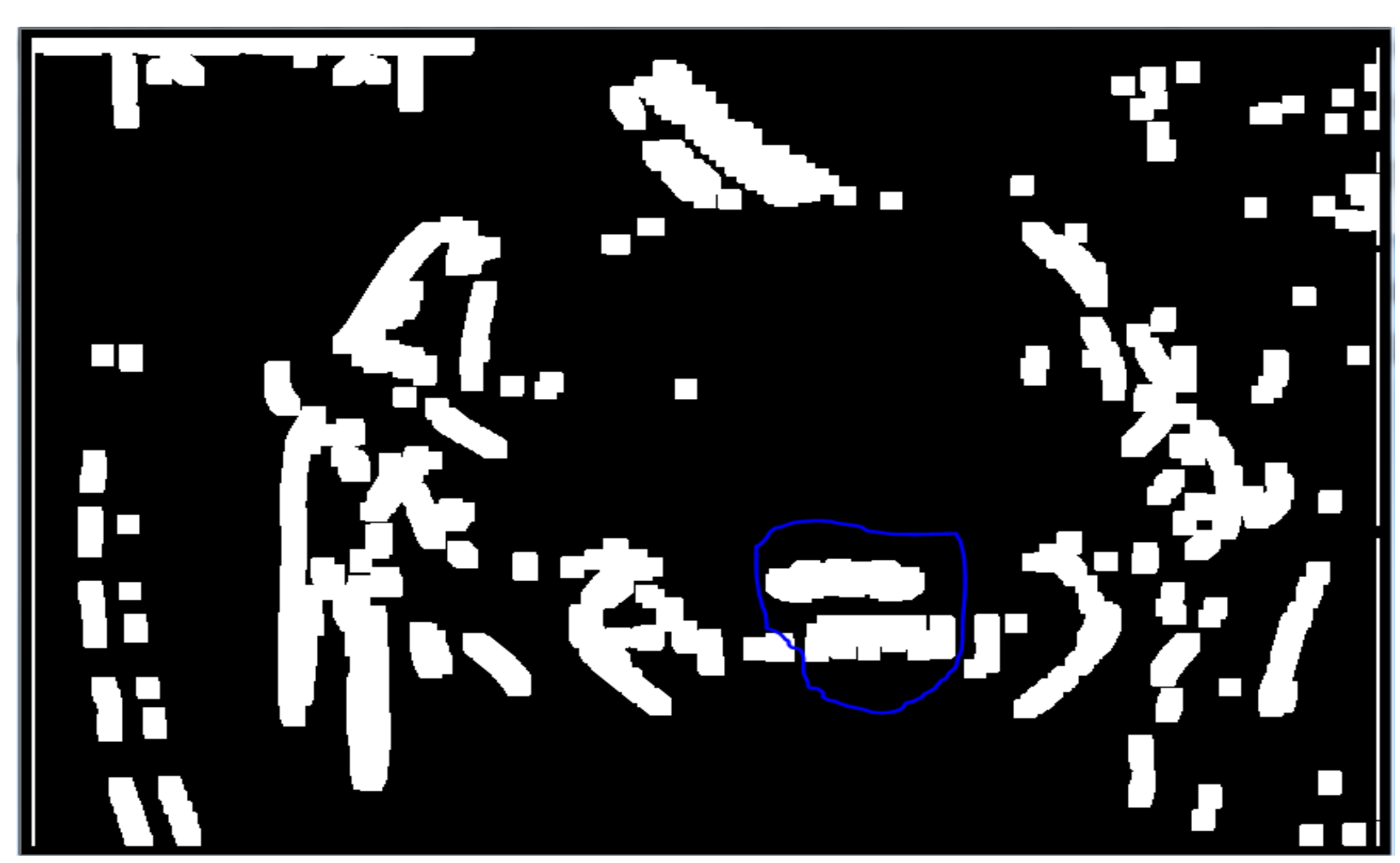

Figure.6Connecting pixels as blocks

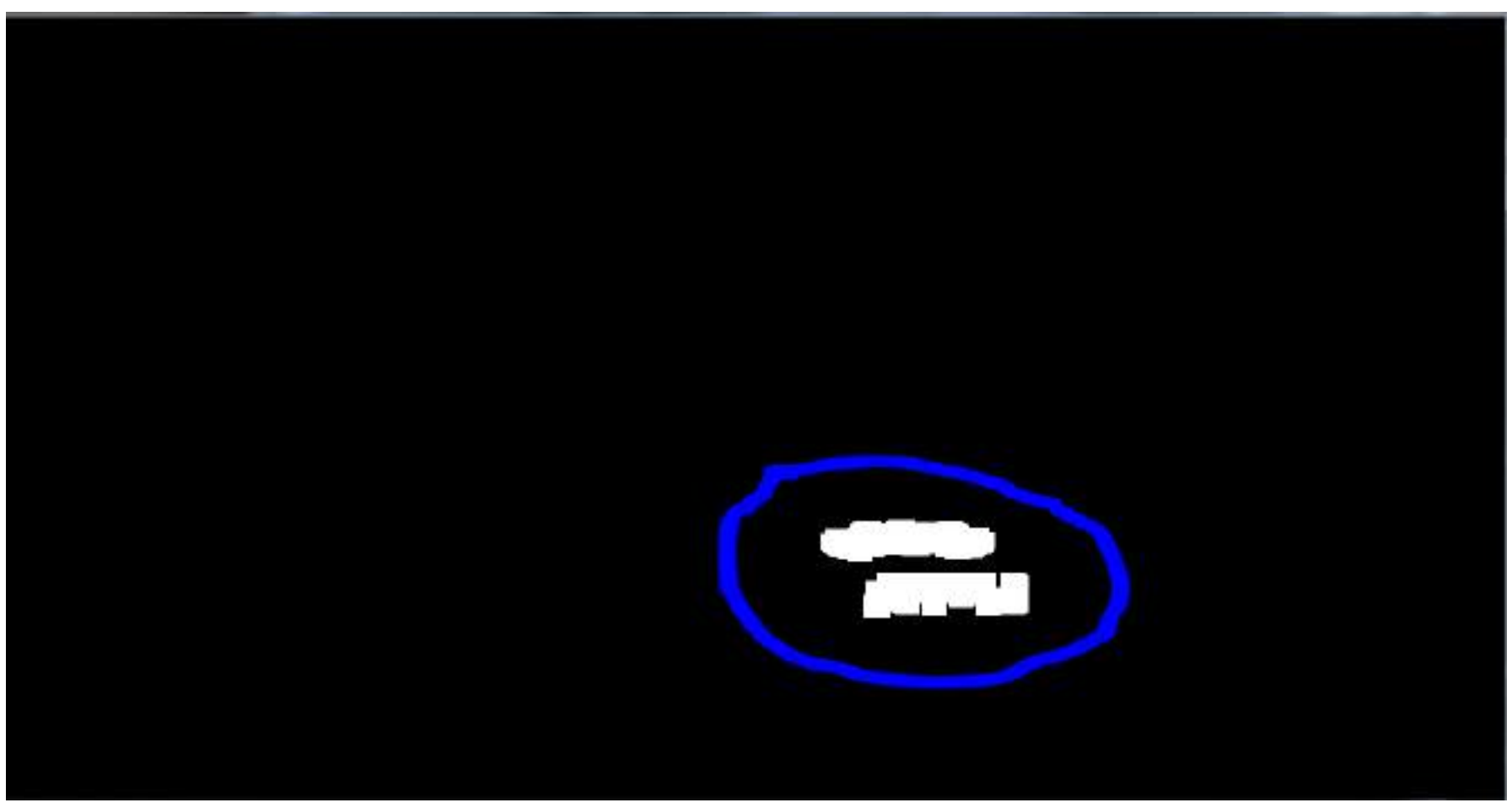

Figure.7Filtration result

### 3.7 EFFECT OF ILLUMINATION

Besides, the lighting in the location that is captured will affect the image processing. While the captured picture is too dark or too bright, it will require to do the balancing process which is to equalize it, optimizing the output of the picture when doing the car plate recognition. For example, if the image is captured in a dark area at night time, the original image may be enhanced by increasing the contrast level, which makes the image become clearer.

### 3.8 WEATHER FACTOR

Last but not least, the weather can affect the image captured by the camera. The image captured will be too blur so that the car plate is not clear. This may cause the car plate recognition system to face difficulty of doing the image processing. For example, if the weather is very cold, leading to the fog formation, the fog will cause the image captured to become blurred.

## IV. DESCRIPTION AND JUSTIFICATION OF THE PROPOSED ALGORITHMS

### 4.1 RGB TO GRAYSCALE CONVERSION ALGORITHM

The original image captured by the camera is in full color format and it consists of 24 bit. This has influences on the application of the image processing. In order to solve the problem, the image can be converted to grayscale from color. This will reduce the image to 8 bit and this will also convert the pixel values to the range from 0 to 255 in which 0 represents black and 255 represents white. By simplifying the value of pixel with RGB to Grayscale conversion algorithm [3], it will help a lot in the calculation of the following image processing stages. An example can be seen below, the RGB image in “Fig.8” after using the algorithm becomes the gray image in “Fig.9”.

Formula: (Red pixel + Blue pixel + Green pixel)/3 = Gray Image Pixel

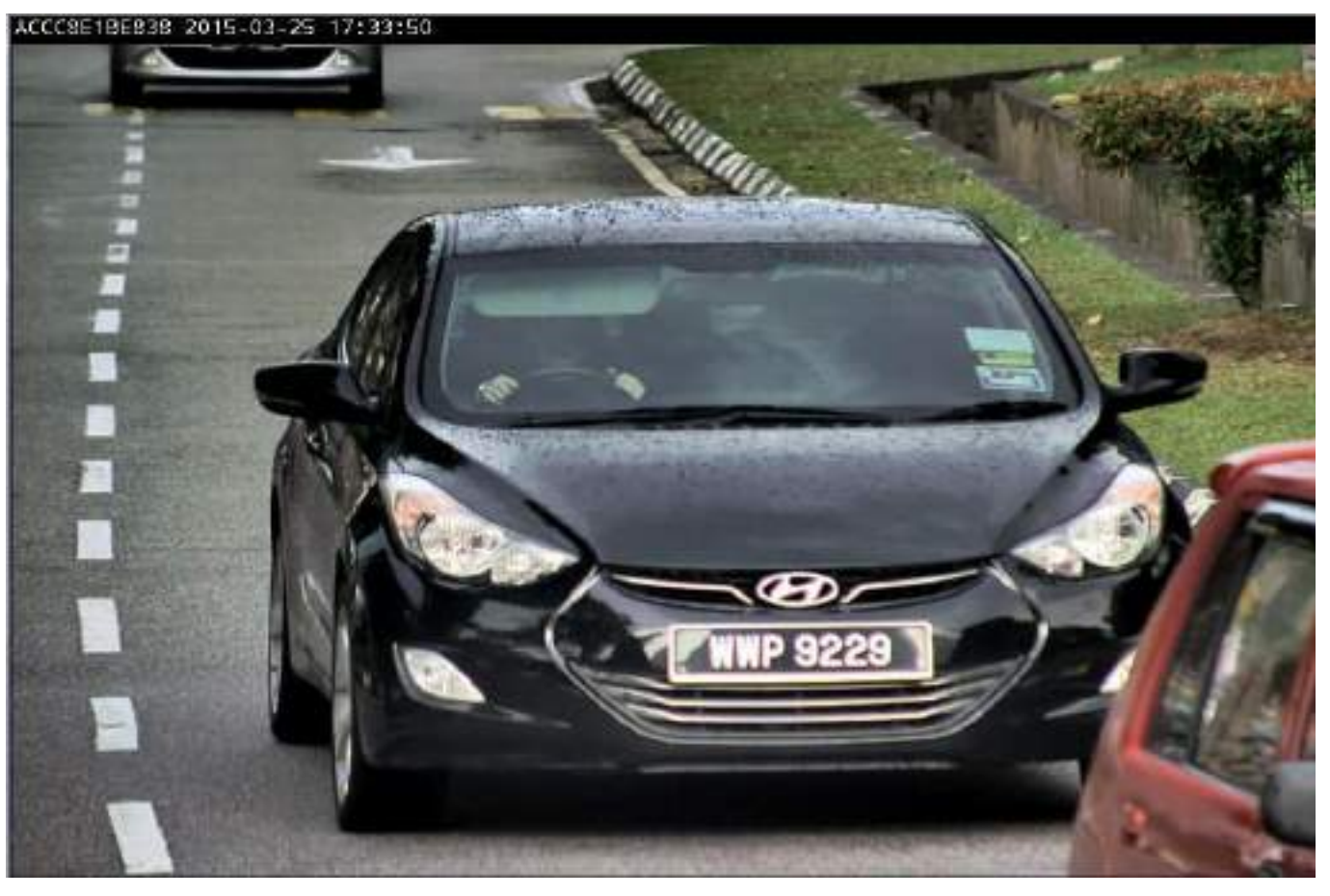

Figure.8RGB Image

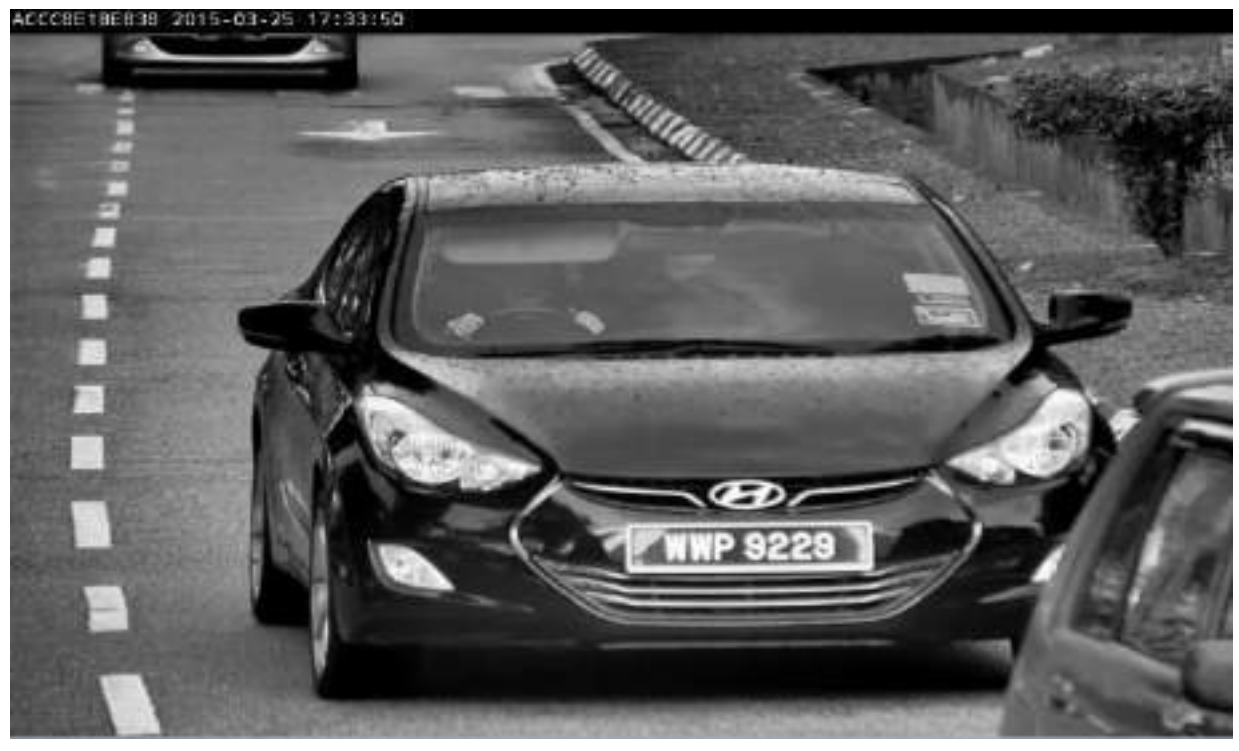

Figure.9 Gray Image

## 4.2 EQUALIZATION ALGORITHM

For solving the problem of unbalanced intensity distribution on the histogram, the equalization algorithm is developed to equalize the intensity distribution of gray image on the histogram so as to enhance the gray image. Through the usage of equalization algorithm, the intensity distribution on the histogram is changed as “Fig.1” shows on the right, compared with the left histogram before equalization. The mathematical theory behind the equalization algorithm is firstly to respectively count the number of each pixel value, secondly to calculate the probability of each pixel value, thirdly to calculate the accumulative probability of each pixel value, at the end to calculate the product of the accumulative probability of each pixel value and 255, returning each product as pixel value of each corresponding pixel in the equalized gray image. For example, after applying the equalization algorithm, the original gray image in “Fig.2” is enhanced as the image in “Fig.12”, which is pretty beneficial to the further processes.

Formula: $E = (G-1)/mn(\sum prob(I(i)))$ , $m$=Height, $n$=Width

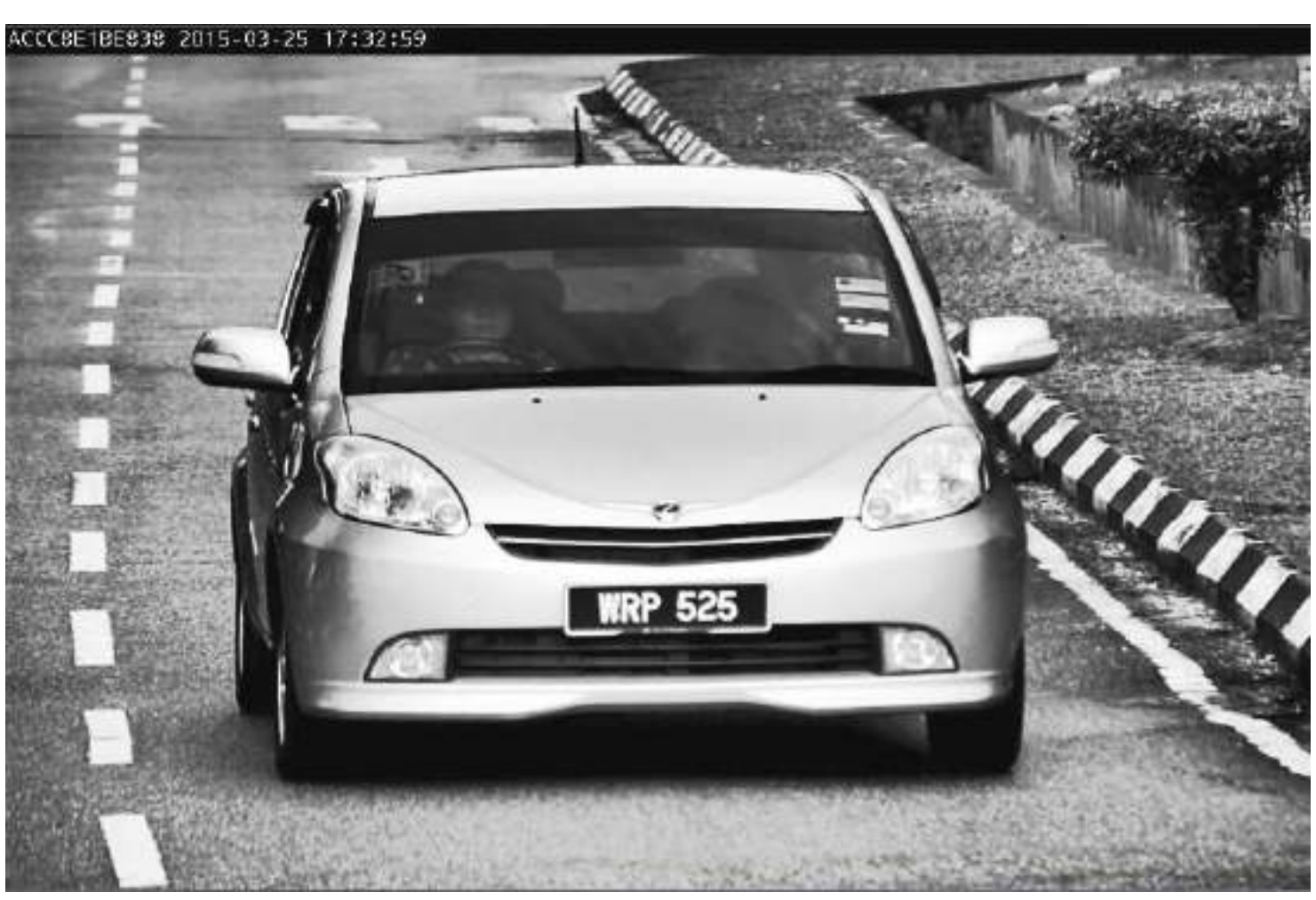

Figure 10.Equalized image

## 4.3 BLUR FILTER ALGORITHM

Blur filter algorithm is an effective method to smooth out the noise of an image. By utilizing the blur filter algorithm, most noises whether in background or foreground of the gray image“Fig.2” are able to be filtered out to produce a smooth gray image “Fig.11” which can be used to produce a better enhancement effect, combined with the equalization algorithm. The mathematical theory behind the blur filter algorithm actually is to use a specific mask to step by step scan a certain number of pixels and calculate the average of the pixels under the mask, returning the average as pixel value to the center pixel under the mask.

## 4.4 EDGE DETECTION ALGORITHM

In order to deal with the problem concerning outline obtainment of plate, the edge detection algorithm is implemented to detect and outline the edge for different objects. Since the plate is an obvious part full of edges, with the help of edge detection algorithm, the developer is able to eliminate lots of objects without edges or apparent edges, only remaining the plate and some other objects with obvious edges as “Fig.12” shows.

The mathematical theory behind the edge detection algorithm technically is to step by step scan a certain number of pixels by using a 3*3 mask, to respectively calculate the average of values of three vertical pixels on the left and the average of values of three vertical pixels on the right, making the center pixel under the mask equal to 255 if the absolute value of the difference of two average is over the specific threshold.

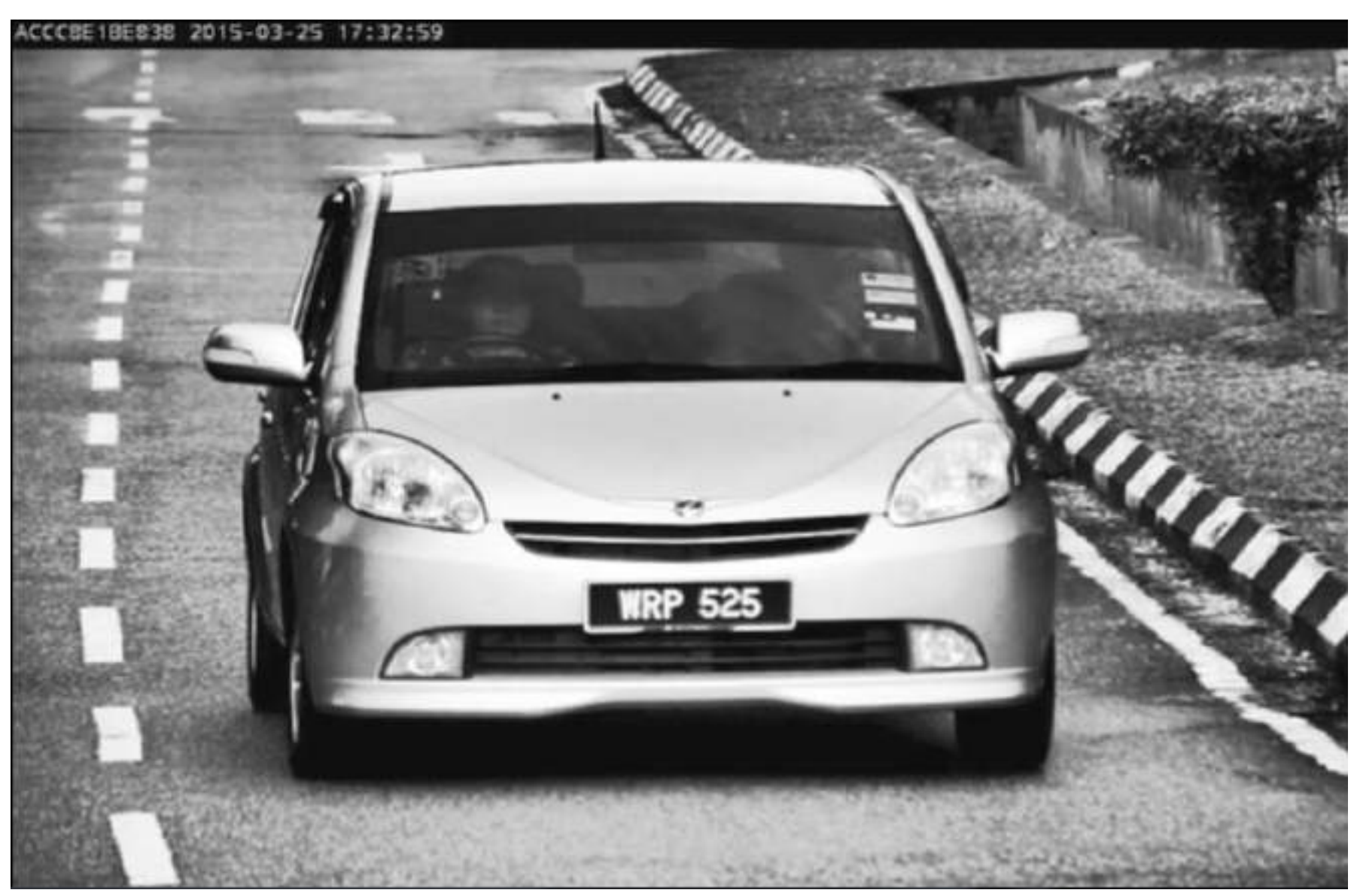

Figure.11 Smooth gray image

Formula:

$$Gx = (z7 + 2z8 + z9)(z1 + 2z2 + z3)$$

$$Gy = (z3 + 2z6 + z9) - (z1 + 2z4 + z7)$$

$$|G| = |Gx| + |Gy|$$

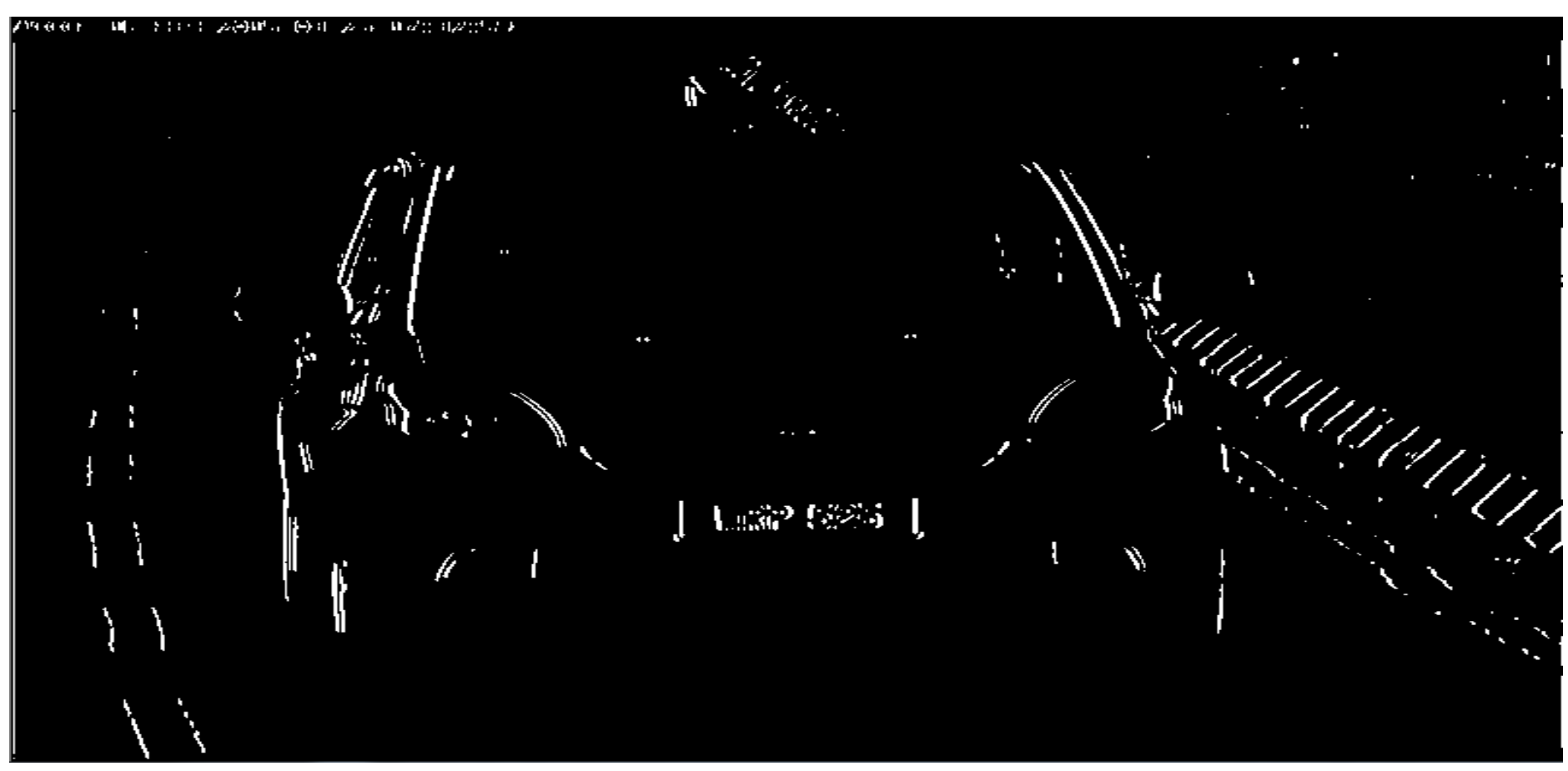

Figure 10. Edge detection

### 4.5 DILATION ALGORITHM

The dilation algorithm is coming out to solve the problem on connection of pixels into blocks. As the meaning of dilation, the algorithm is to dilate the relevant scattered pixels to connect them together so as to make each object a complete block. For instance, after using dilation algorithm, the image as “Fig.16” is changed into the image as “Fig.13”. The mathematical theory behind the dilation algorithm is to step by step scan a certain number of pixels using a 3*3 mask, to identify whether there is a pixel value equal to 255 among the 8 pixels surrounding the center pixel under the mask, if yes, making the pixel value of the center pixel equal to 255.

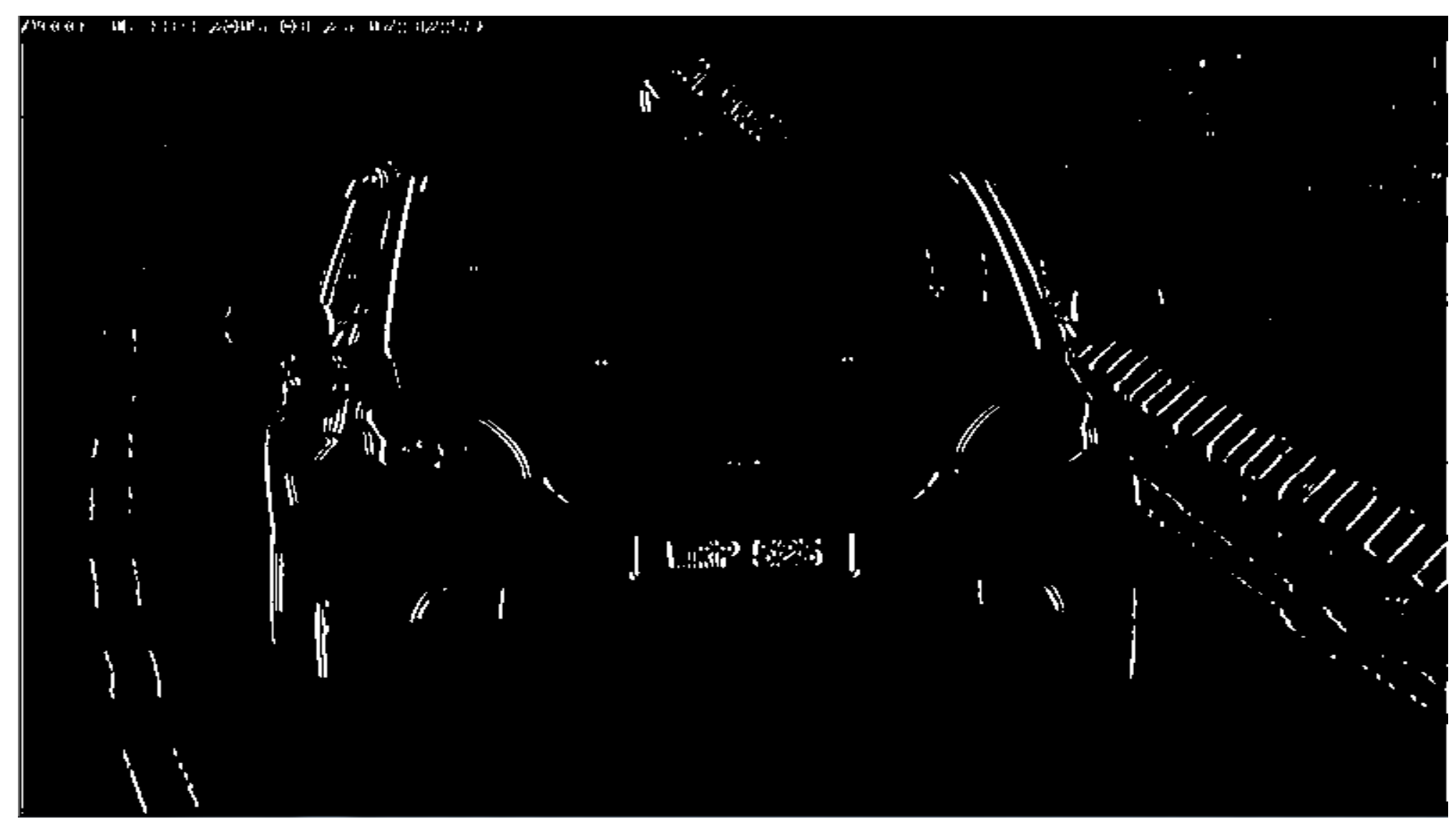

Figure.13 Scattered Pixels

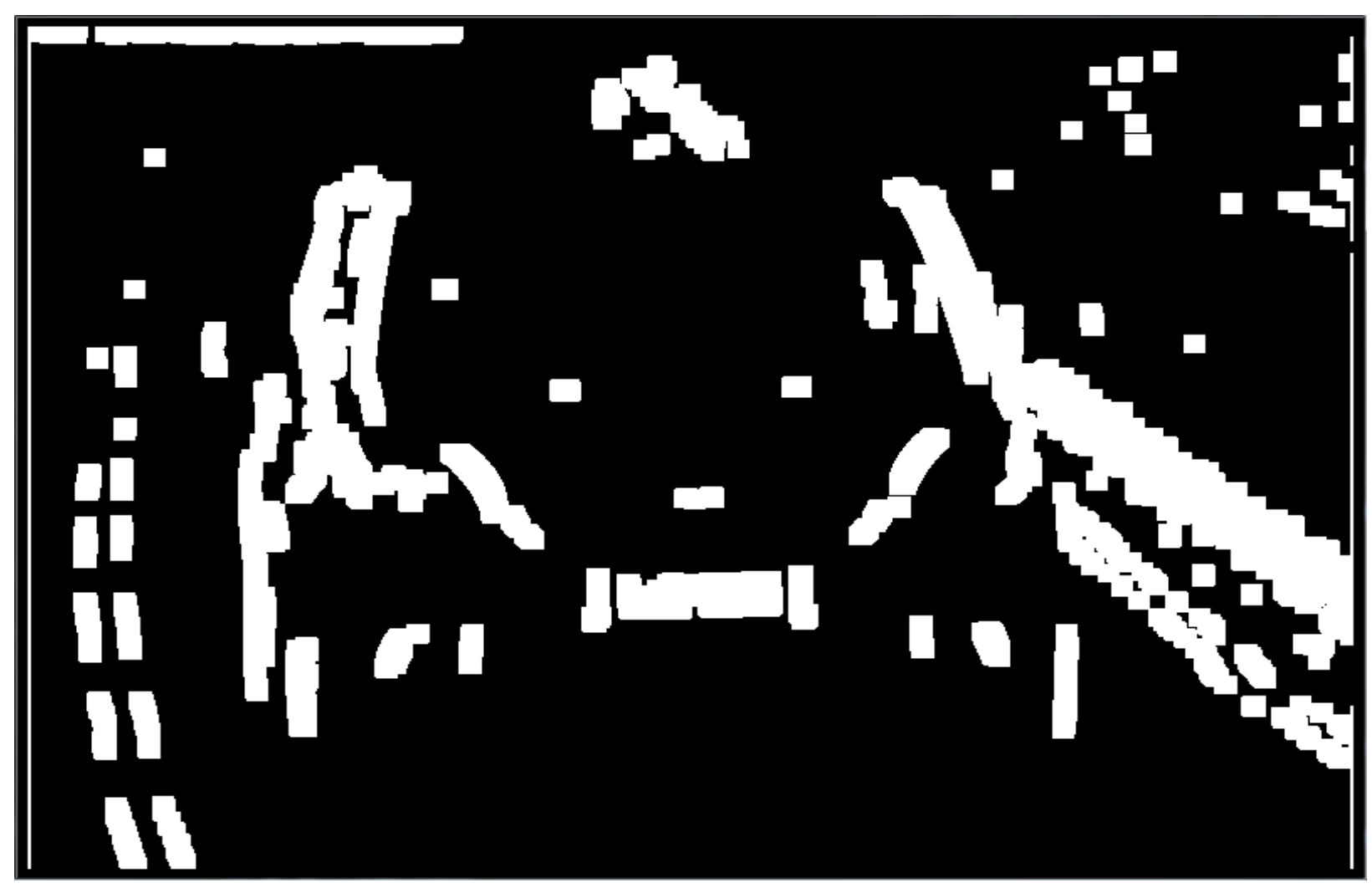

Figure.14 Complete blocks

## 4.6 SEGMENTATION ALGORITHM

To deal with the problem concerning individual processing of each block, the find-contours function from Open CV is utilized in the segmentation algorithm. After using the segmentation algorithm, the image is changed from "Fig.14" to "Fig.15", which successfully segment various blocks in one image as different individuals labeled with different colors.

The mathematical theory behind the segmentation algorithm is to firstly input the whole image into the find-contours function from Open CV, secondly store all the outputs of contour into a vector, and thirdly according to the vector, draw each contour in random color depending on the clock of CPU.

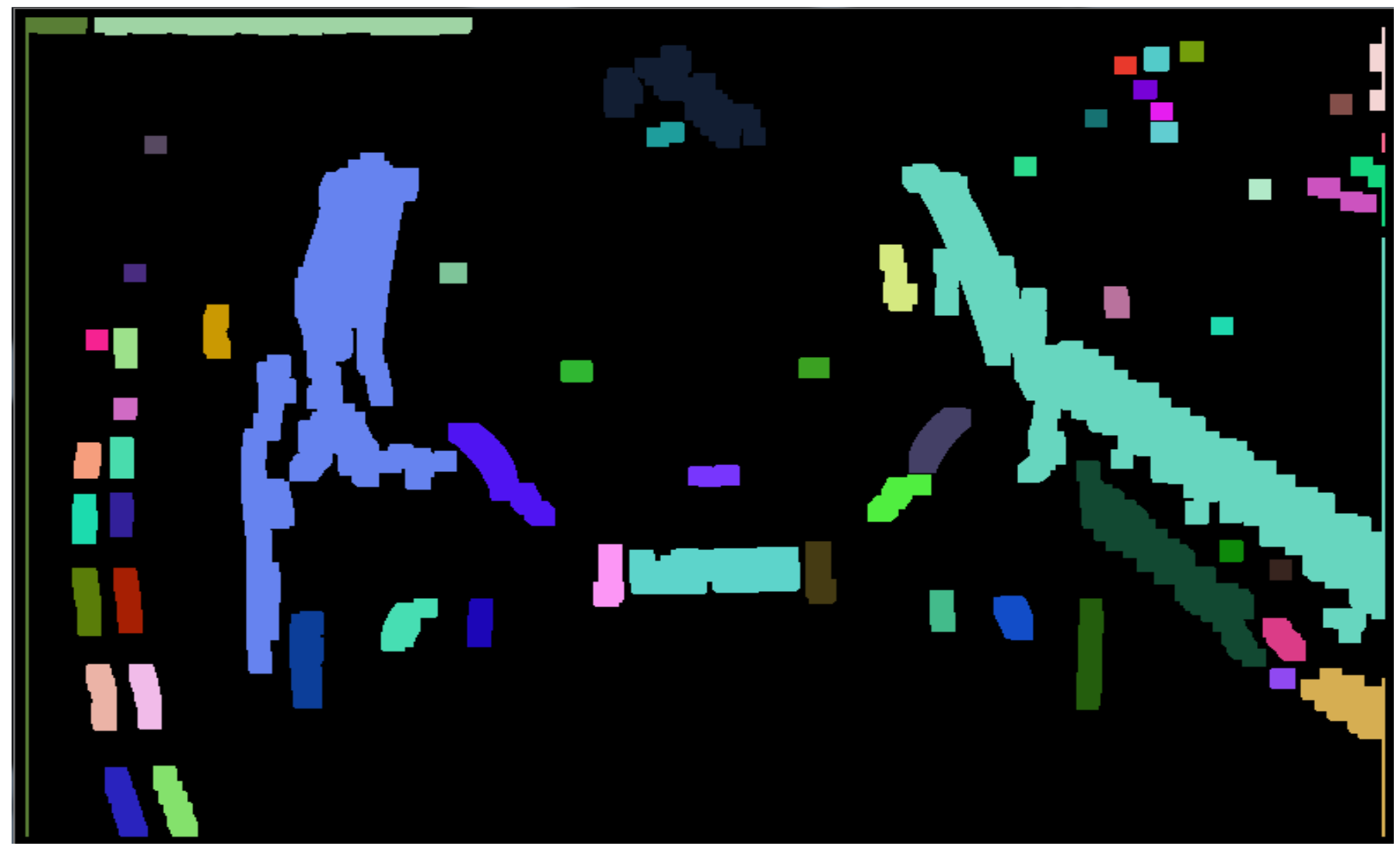

Figure.15 After Segmentation

## 4.7 NOISE REMOVAL ALGORITHM

After the processing of segmentation, in order to remove all the irrelevant noise objects to get only the plate left eventually, the noise removal algorithm is coming out as the solution to achieve the purpose. At the beginning, the noise removal algorithm is developed by applying the heuristic features since it is enough to filter out all the noise objects in lots of images after the segmentation to leave the plate only. However, through many experiments, it is found that the algorithm applying the heuristic features is unable to process some images successfully to leave the plate only. Therefore, by doing a lot of researches, the developer finally is determined to utilize the new technique which is proposed by [1] into the noise removal algorithm. In fact, the developer is not just direct to fully use the original method proposed by [1] but actually to modify and optimize it by combining with the area feature as a new noise removal algorithm also shown in "Fig.16 & Fig.17".

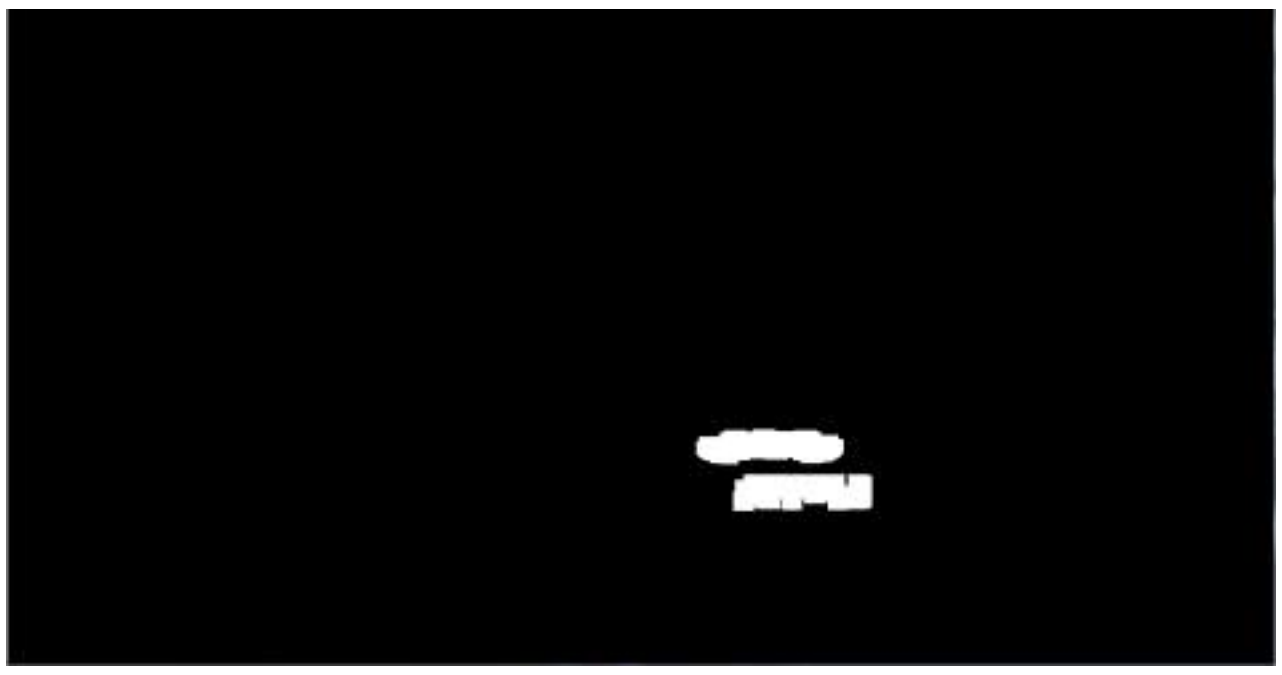

Figure.16 Heuristic Feature Results

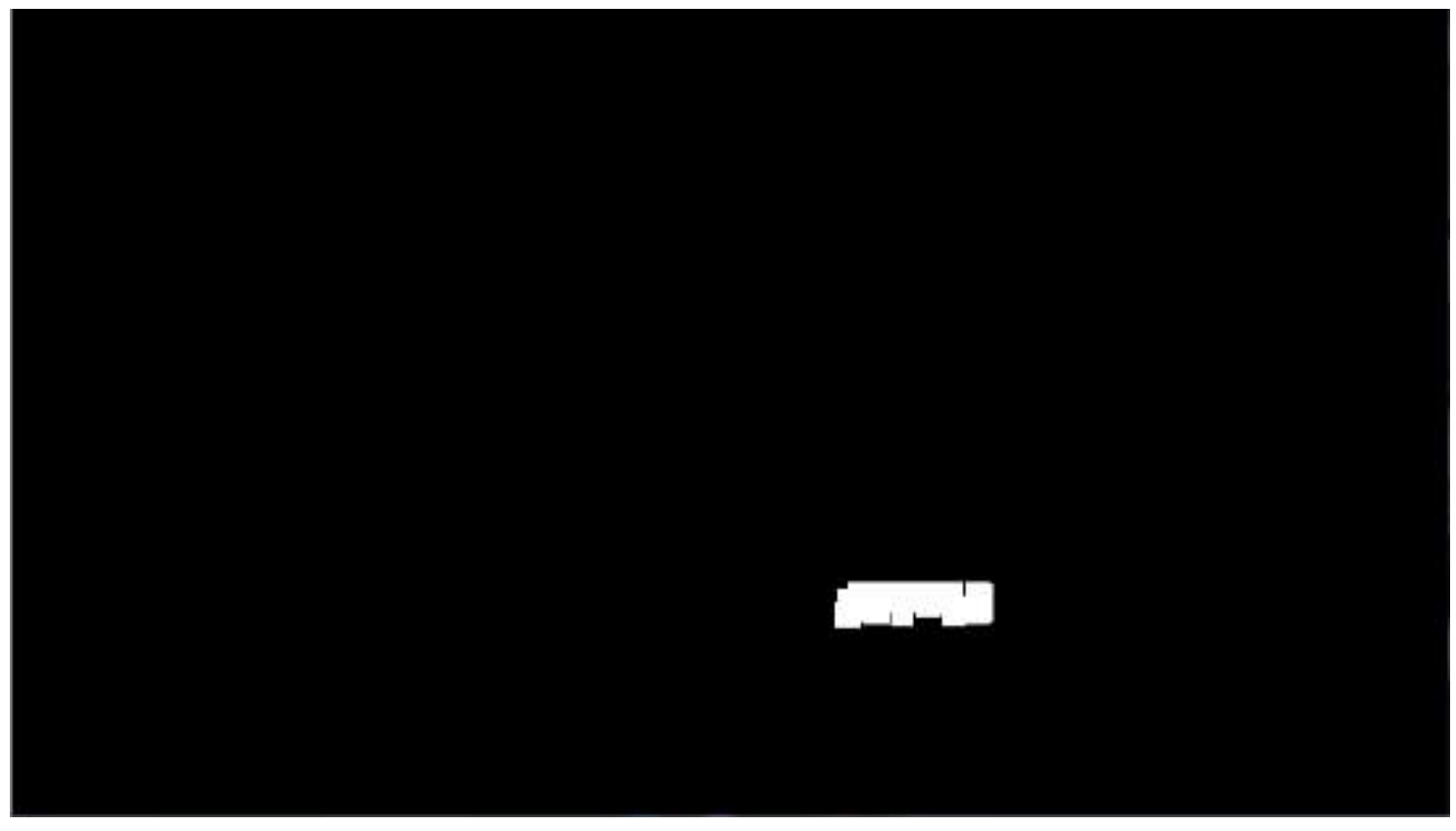

Figure.17 New Algorithm Result

## 4.7 PLATE EXTRACTION ALGORITHM

After the segmentation and noise removal, the ideal rectangle box has been selected and the car plate position has been found. Now the car plate can be captured from the final processed image. There is an example of the captured plate as "Fig.18" shows which is captured successfully by applying the plate extraction algorithm.

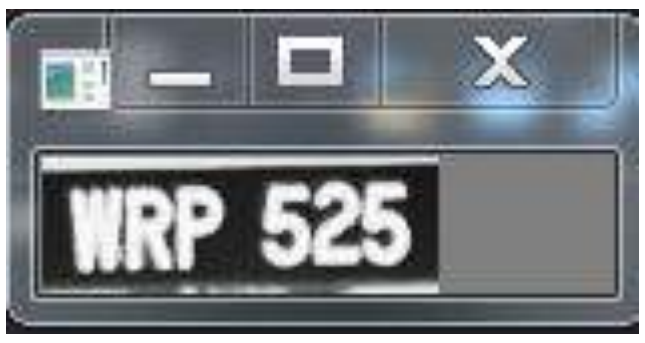

Figure.18 Captured Plate

## V. CRITICAL COMMENTS AND ANALYSIS

The setting of the threshold in edge detection and the mask dimension in blur filter and dilation is pretty of significance to the following processes. For example, if the setting of the mask dimension in dilation is not appropriate, it may influence the accuracy of locating the plate so that the complete region of the plate fails to be located and at the end it comes out with an incomplete plate extraction like “Fig.19” compared with the original plate in “Fig.20”. Actually, through multiple experiments, it is found that the reason for this problem is due to the inappropriate setting of the mask dimension in dilation, which causes the pixels of the plate to fail to connect together completely. Therefore, the setting of the relevant parameters like threshold, mask dimension and so on has to be focused on by the developer so as to achieve a good result.

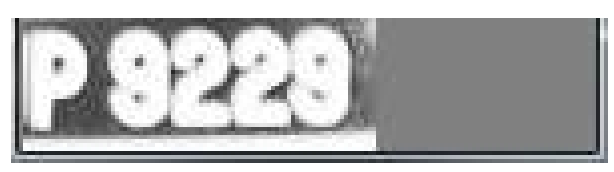

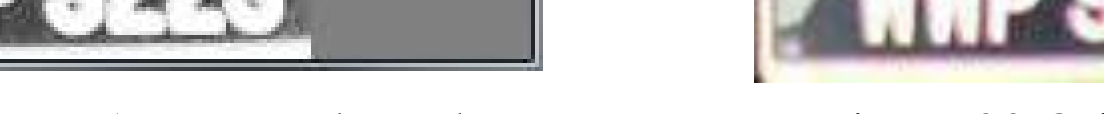

Figure.19 Incomplete Plate

Figure.20 Original Plate

Other than that, the environmental factors like lighting and weather also have a lot of influences to the processing results, since the environmental is direct to affect the quality of the original images captured by the camera. Therefore, to avoid this as far as possible, the camera device is suggested to be placed in the location with comparatively stable and sufficient light, and equipped with weather-resistant measures.

## VI.CONCLUSION

During the development of the project, not only did the developer acquire the domain knowledge concerning the license plate recognition but also has learnt more C++ programming skills and features, which impressed the developer a lot. To make a long story short, the license plate recognition system project has been achieved by developing and utilizing the proposed algorithms. The goal of the project is to achieve the plate detection, which means to locate and extract the plate from the image at the end by applying various proposed methods in order to remove all the noises of the image. The license plate recognition system will help to reach the objectives which are to enhance the car plate recognition, to assist police in finding the stolen cars and to limit down the speeding cars.

## VII. ACKNOWLEDGMENT

The authors would like to share gratitude to Mr Umapathy Eaganathan, Lecturer in Computing, Asia Pacific University, Malaysia also Miss Angel Aron for the constant support and motivation helped us to participate in this International Conference and also for journal publication.